%% file: acl_latex.tex
\pdfoutput=1

\documentclass[11pt]{article}

\usepackage{authblk}

\usepackage[]{acl}

\usepackage{times}
\usepackage{latexsym}

\usepackage[T1]{fontenc}

\usepackage[utf8]{inputenc}

\usepackage{microtype}

\title{On Text-based Personality Computing: Challenges and Future Directions}

\author[1]{Qixiang Fang}
\author[1]{Anastasia Giachanou}
\author[1]{Ayoub Bagheri}
\author[1]{Laura Boeschoten}
\author[1]{\\Erik-Jan van Kesteren}
\author[1]{Mahdi Shafiee Kamalabad}
\author[1, 2]{Daniel L. Oberski}
\affil[1]{Utrecht University, Utrecht, the Netherlands}
\affil[2]{University Medical Center Utrecht, Utrecht, the Netherlands}

\affil[ ]{\texttt{\{q.fang, a.giachanou, a.bagheri, l.boeschoten\}@uu.nl}}
\affil[ ]{\texttt{\{e.vankesteren1, m.shafieekamalabad, d.l.oberski\}@uu.nl}}

\begin{document}
\maketitle
\begin{abstract}
Text-based personality computing (TPC) has gained many research interests in NLP. In this paper, we describe 15 challenges that we consider deserving the attention of the NLP research community. These challenges are organized by the following topics: personality taxonomies, measurement quality, datasets, performance evaluation, modelling choices, as well as ethics and fairness. When addressing each challenge, not only do we combine perspectives from both NLP and social sciences, but also offer concrete suggestions. We hope to inspire more valid and reliable TPC research.
\end{abstract}

\input{writing/1Introduction}

\input{writing/3Background}
\input{writing/4Taxonomies}
\input{writing/5Measurement}
\input{writing/6Datasets}

\input{writing/7Evaluation}

\input{writing/8Models}

\input{writing/9Theory}
\input{writing/10Ethics}

\input{writing/11Conclusion}
\bibliography{bibliography}
\bibliographystyle{acl_natbib}

\appendix

\begin{table*}[htb]
\centering
\small
\begin{tabular}{llllll}
\hline
\textbf{Source} & \textbf{Name} & \textbf{Size} & \textbf{Taxonomy} & \textbf{Text Type} & \textbf{Language}\\
\hline
\citet{barriere_wassa_2022} & WASSA 2022 & 2655 & Big-5 & News and essays & en\\
\citet{bassignana2020personal} & Personal-ITY & 1048 & MBTI & YouTube comments & it\\
\citet{biel2012youtube} & Youtube  & 442 & Big-5 & YouTube vlogs  & en\\ & Personality & & & transcripts\\
\citet{gjurkovic_pandora_2021} & PANDORA$^*$ & 10288 & Big-5 \&  & Reddit posts  & en, es, fr, it, de, pt\\ & & & MBTI& & nl, eo, sv, pl\\ 
\citet{gjurkovic_reddit_2018} & Reddit9k$^*$ & 9111 & MBTI & Reddit posts  & en \\
\citet{jansen_introducing_2020} & MULAI$^*$ & 26 & Big-5  & Conversation  & en\\ & & & &transcripts\\
\citet{koutsombogera_modeling_2018} & MULTISIMO$^*$ & 49 & Big-5 & Conversation  & en\\ & & & &transcripts\\
\citet{kaggleData} & Kaggle & 8600 & MBTI & Forum posts  & en\\
\citet{pennebaker1999linguistic} & Essays & 2479 & Big-5 & Essays & en\\
\citet{plank-hovy-2015-personality} & Twitter & 1500 & MBTI & Twitter posts & en\\
\citet{ponce2016chalearn} & ChaLearn V1 & 10000 & Big-5 & YouTube vlogs  & en\\ & & & &transcripts\\
\citet{ramos_building_2018} & b5$^*$ & 1082 & Big-5 & Facebook status,  & pt\\ & & & & referring expressions \\ & & & & and scene descriptions\\
\citet{rangel2015overview} & PAN 2015$^*$ & 300 & Big-5 & Twitter posts & en, es, it, nl\\
\citet{tolins_multimodal_2016} & Personality & 6 & Big-5 & Conversation &  en\\ & Dyads$^*$ & & & transcriptions\\
\citet{tolins_verbal_2016} & Storytron & 44 & Big-5 & Conversation &  en\\ & Retellings$^*$ & & & transcriptions\\
\citet{verhoeven_clips_2014} & CSI & 697 & Big-5 \& & Essays and reviews & nl\\ & & & MBTI \\
\citet{verhoeven_twisty_2016} & TWISTY & 18168 & MBTI & Twitter posts & de, it, nl, fr, pt, es\\ 
\citet{wen_automatically_2021} & PELD & 711 & Big-5 & TV show dialogs & en\\
\hline
\end{tabular}
\caption{\label{table:datasets}
Overview of Shareable Datasets. *Data available upon simple request (e.g., via a form on the dataset hosting website).
Abbreviations of languages are based on ISO 639-1 Code.
}
\end{table*}

\section{Literature Search and Overview}
\label{sec:appendixA}
To search for relevant TPC studies in the ACL Anthology, we use the following search terms: “personality”, “personality prediction”, “personality computing” and “personality recognition”. The search date is August 29, 2022. We also sort the results by “relevance” and “year of publication”, separately, as these two options return different search results. For all the returned papers, we check their titles, abstracts and (when necessary) main texts, and retain those that fall under one of the following categories:

\paragraph{Dataset Type-A}
Papers focusing on introducing new datasets for TPC. Some also benchmark TPC models against new datasets. There are 7: \citet{luyckx_personae_2008, verhoeven_clips_2014, plank-hovy-2015-personality, verhoeven_twisty_2016, bassignana_matching_2020, gjurkovic_pandora_2021, barriere_wassa_2022}.

\paragraph{Dataset Type-B}
Papers that introduce datasets for non-TPC goals (e.g., multimodal PC, natural language generation, modelling laughter in conversations) but happen to include personality and text data that can be used for TPC. There are 5: \citet{tolins_verbal_2016, tolins_multimodal_2016, koutsombogera_modeling_2018, ramos_building_2018, jansen_introducing_2020}.

\paragraph{TPC as End}
Empirical papers where TPC is the main goal. There are 31: \citet{mairesse_automatic_2006, resnik_using_2013, sinha_mining_2015, fung_zara_2016, kamijo-etal-2016-personality, levitan_identifying_2016, liu_recurrent_2016, das_developing_2017, liu_language-independent_2017, siddique_zara_2017, gjurkovic_reddit_2018, kampman_investigating_2018, tighe_modeling_2018, vu_lexical-semantic_2018, zamani_predicting_2018, pizzolli_personality_2019, yamada_incorporating_2019, cornelisse_inferring_2020, iwai_development_2020, lynn_hierarchical_2020, suman_multi-modal_2020,  culnan_me_2021, hull_personality_2021, kishima_construction_2021,  v_ganesan_empirical_2021, yang_learning_2021, yang_psycholinguistic_2021,  ghosh_team_2022, kerz_pushing_2022, kreuter_items_2022, li_prompt-based_2022}.

\paragraph{TPC as Means}
Empirical papers where TPC is not the main goal but used for further purposes like emotion detection, natural language generation and sociolinguistic analysis. There are 17: \citet{mairesse_personage_2007, mairesse_trainable_2008, makatchev_perception_2011, gill_perceptions_2012, roshchina_evaluating_2012, preotiuc-pietro_role_2015, yang_using_2015, guntuku_current_2018, lan_definite_2018, al_khatib_exploiting_2020, iwai_acquiring_2020, he_personality_2021, hosseinia_usefulness_2021, uban_understanding_2021, vettigli_empna_2021, wen_automatically_2021, saha_stylistic_2022}.

\vspace{2mm}
Note that we exclude papers that mean PC as the computing of personality disorders or personal profiles. In total, we find and review 60 TPC papers. Also, see Appendix~\ref{sec:appendixB} for 18 shareable TPC datasets that we find via these 60 papers. 

\section{Overview of Shareable TPC Datasets}
\label{sec:appendixB}
See Table \ref{table:datasets}.

\end{document}

%% file: writing/1Introduction.tex
\section{Introduction}

According to the APA Dictionary of Psychology~\citep{apaDictionary23}, personality refers to personality traits, which are “relatively stable, consistent, and enduring internal characteristics inferred from a pattern of behaviours, attitudes, feelings and habits in individuals”. Knowledge about personality can be useful in many societal and scientific applications. For instance, it can help individuals choose learning styles~\citep{komarraju2011big} and occupations~\citep{kern2019social} suited for their personality; it can help clinical psychologists to better understand psychological disorders~\citep{khan2005personality} and to deliver personalised treatment plans for mental health patients~\citep{bagby2016role}; changes in personality can even help with early diagnosis of Alzheimer’s~\citep{robins2011personality} and Parkinson’s disease~\citep{santangelo2017personality}.

Traditionally, personality assessment is based on self- and other-report questionnaires, which is labour- and time-intensive. Recently, however, automatic personality assessment based on user-generated data (e.g., texts, images, videos) and machine learning algorithms has become a popular alternative. This is known as \textbf{personality computing (PC)}~\citep{phan2021personality}, among many other names\footnote{Personality recognition~\citep{liu_language-independent_2017}, personality identification~\citep{das_developing_2017}, personality detection~\citep{yang_learning_2021} and personality prediction~\citep{yamada_incorporating_2019}}.


In this paper, we focus on evaluating PC research in NLP, where personality is primarily inferred from text, such as tweets and Reddit posts~\citep{hosseinia_usefulness_2021}, conversations~\citep{mairesse_automatic_2006} and speech transcriptions~\citep{das_developing_2017}. We refer to such research as \textbf{text-based personality computing (TPC)}. 


In TPC, on the one hand, we see an increasing number of datasets curated, complex deep-learning algorithms adopted, and (sometimes) high prediction scores achieved. On the other hand, we see relatively little discussion about open challenges and future research directions.
For example, the presence of measurement error in questionnaire-based personality scores remains an un(der)addressed issue. Given that such scores are often used as the gold standard for training and validating TPC algorithms, we find it important to discuss related implications and remedies. Another relevant issue concerns how to reduce the risks of TPC research.


Therefore, in this paper, we reflect on current TPC research practices, identify open challenges, and suggest better ways forward. To spot such challenges, we conduct a literature search in the scope of the ACL Anthology\footnote{https://aclanthology.org/}. While there are TPC papers published in other venues, we consider our selection from ACL Anthology a good representation of TPC research in NLP. Appendix~\ref{sec:appendixA} describes our search strategy and results in detail. In total, we find and review 60 empirical TPC papers, based on which we identify 15 challenges. They are organized by the following topics: personality taxonomies (\S\ref{sec_taxonomies}), measurement quality (\S\ref{sec_measurements}), datasets (\S\ref{sec_datasets}), performance evaluation (\S\ref{sec_evaluation}), modelling choices (\S\ref{sec_modelling}), as well as ethics and fairness (\S\ref{sec_ethics}). We discuss each challenge and give concrete suggestions where we also draw on broader NLP and social science literature. 
Furthermore, during our literature review, we identify 18 TPC datasets that can be (re)used by other researchers. We summarize them in Appendix~\ref{sec:appendixB}.

Note that our paper focuses on identifying and discussing TPC-related research challenges instead of providing a comprehensive overview of past TPC studies. For the latter, we refer you to the survey papers by \citet{stajner_survey_2020, FeiziDerakhshi2021TextbasedAP, mushtaq2022text}.

%% file: writing/3Background.tex
\section{Background}
\subsection{Current TPC in a Nutshell}
TPC concerns computing personality information from texts. This can be either a regression or a classification task, depending on whether the personality measurements are continuous or discrete. Supervised learning has been the predominant approach, relying on text datasets labelled with personality traits via either self-report or crowdsourced annotations. Among many others, log-linear models~\citep{Volkova2015InferringLU}, random forests~\citep{levitan_identifying_2016}, GloVe embeddings with Gaussian processes~\citep{Arnoux201725TT}, recurrent neural networks~\citep{liu_language-independent_2017}, convolutional neural networks~\citep{Majumder2017DeepLD}, support vector machines~\citep{lan_definite_2018}, ridge regression~\citep{he_personality_2021}, graphical networks~\citep{yang_psycholinguistic_2021} and transformers~\citep{kreuter_items_2022} have been used. Another popular, psycho-linguistically motivated approach are dictionaries/lexicons~\citep[e.g.,][]{oberlander2006whose, sinha_mining_2015, das_developing_2017}. They are typically lists of curated terms that have pre-assigned weights associated with different personality traits. This allows researchers to compute personality scores from texts by simply matching a dictionary to the texts and aggregating the weights of matched words and phrases.

\subsection{Related Work}
\label{sec:relatedWork}
We find three relevant TPC survey papers. The first one, \citet{stajner_survey_2020}, not only summarizes previous research in TPC but also discusses three interesting issues: the difference between MBTI and the Big-5 (two most popular personality taxonomies, see \S\ref{sec_taxonomies}), the difficulty in predicting MBTI from Twitter data, and ethical concerns about TPC. We find all three discussions necessary and helpful. Our paper not only provides our own (differing) perspectives on these issues, but also raises and discusses many others. The other two survey papers, \citet{mushtaq2022text} and \citet{FeiziDerakhshi2021TextbasedAP}, after reviewing prior TPC research, very briefly suggest some open challenges in TPC research. While the open challenges mentioned in these two papers (e.g., better data quality, data sharing, and ethics) partially overlap with our list of challenges, we adopt a more evidence-based approach to listing challenges, engage in much more thorough discussion, and offer concrete solutions. Furthermore, our list of challenges goes beyond those in these two papers (e.g., measurement error reduction, performance expectation, joint personality modelling).

We also find three additional papers~\citep{bleidorn2019using, stachl2020personality, phan2021personality} concerning issues in general PC research. Our paper, in contrast, focuses on challenges specific to TPC research. 

Lastly, TPC is closely related to other fields, such as automatic emotion recognition~\citep{barriere_wassa_2022}, opinion mining~\citep{hosseinia_usefulness_2021} and mental health prediction~\citep{guntuku_current_2018}. 
These fields are all concerned with the computation of social science constructs (see \S\ref{sec_measurements}).

%% file: writing/4Taxonomies.tex
\section{Personality Taxonomies}
\label{sec_taxonomies}
A TPC research project typically starts by choosing a personality taxonomy, which is a descriptive framework for personality traits~\citep{john1999big}.  Among the 60 papers we review, we find two prominent taxonomies: the Myers-Briggs Type Indicator (MBTI; 14 papers) and the Big-5 (45 papers).\footnote{Also, Enneagram and HEXACO each appeared once.} Fifty of these papers adopt \emph{either} the MBTI \emph{or} the Big-5 but not both. This invites the first challenge:

\begin{quote}
    \label{C1}
    \textbf{Challenge 1 (C1): MBTI vs. Big-5}
\end{quote}

The MBTI originated from the theoretical work of~\citet{jung1971psychological} and was further developed by~\citet{briggs1995gifts}. It proposes four personality dimensions that characterize people’s differences in perception and judgement processes: Extraversion/Introversion (E/I), Sensing/iNtuition (S/N), Thinking/Feeling (T/F), and Judgement/Perception (J/P). Individuals are classified into one of the two possible categories across each dimension (e.g., INFP and ESTJ). 


In contrast, the Big-5 was developed based on the lexical hypothesis: \emph{all important personality traits must have been encoded in natural language and therefore, analysis of personality-related terms should reveal the true personality taxonomy}~\citep{goldberg1990alternative}. Independent research groups~\citep{cattell1946description,goldberg1982ace,costa1992professional,tupes1992recurrent} investigated this hypothesis. They identified numerous English terms that might describe inter-individual differences (e.g., warm, curious), asked participants to rate how well these terms describe them on numerical scales, and factor-analyzed the responses, which revealed five consistent dimensions of personality: Openness (O), Conscientiousness (C), Extraversion (E), Agreeableness (A), and Neuroticism (N).\footnote{The Big-5 is also called OCEAN.} Furthermore, each dimension includes six finer-grained sub-traits called \textit{facets} (e.g., agreeableness includes trust, straightforwardness, altruism, compliance, modesty and tender-mindedness). Big-5 is the most widely accepted and researched taxonomy of personality traits in psychology (as opposed to the popularity of MBTI in non-academic settings like job interviews)~\citep{phan2021personality}. 

We recommend the Big-5 over MBTI for the following reasons:

First, the Big-5 is a more realistic and accurate personality taxonomy. It scores individuals along a continuous spectrum, which describes inter-individual differences more accurately and preserves more information (as opposed to MBTI’s dichotomous approach). Also, the Big-5 includes facets, which allows for finer-grained analysis of personality. Facets are also more predictive of life outcomes (compared to dimensions; \citealt{mershon1988number, paunonen2001big}). A potential new TPC research direction can be to predict facets in addition to dimensions. 

Second, Big-5 has a much stronger empirical basis than MBTI. Namely, it is grounded in large-scale quantitative analysis of natural language and survey data. Also, Big-5 questionnaires have undergone much more extensive development and validation processes than MBTI's. Consequently, many validated Big-5 questionnaires exist, which vary in length (15-240 items), the inclusion of facet measures, as well as the target populations (e.g., nationalities, professions, languages, and age groups).\footnote{E.g., the NEO Personality Inventory~\citep{mccrae1989neo}, the Revised NEO-PI and the NEO Five Factor Inventory~\citep{costa1992professional}, the Big-5 Inventory (BFI)~\citep{benet1998cinco}, BFI-2~\citep{soto2017next}, the Short 15-item Big-5 Inventory~\citep{lang2011short}.} This enables researchers to choose a questionnaire most appropriate given a population of interest, available resources (e.g., can you afford a longer questionnaire?) and research interests (e.g., are you interested in facets?). In comparison, MBTI is purely theory-driven, lacks empirical support and has officially only four questionnaires that have not been thoroughly tested~\citep{pittenger1993measuring, nowack1996myers,grant2013goodbye}.\footnote{Form M with 93 items, Form M self-scorable with 93 items, Form Q with 93 items and Step III Form with 222 items. See https://www.myersbriggs.org/.} Therefore, compared to MBTI, Big-5 is a much more credible and flexible choice for research purposes.  


Third, the Big-5 is rooted in natural language (i.e., the lexical hypothesis), suggesting that Big-5-related cues may be more present than do MBTI-related cues in text data. This conjecture is supported by \citet{stajner_why_2021}, who find either insufficient or mixed signals for MBTI dimensions in tweets and short essays. 

Nevertheless, despite the many advantages of Big-5 over MBTI, we acknowledge that studying MBTI can still be useful given its popularity in non-academic settings~\citep{lloyd2012myers}. Furthermore, it is also important to mention that Big-5 is not without criticisms and other personality taxonomies may be preferred, which invites the next challenge. 


\begin{quote}
    \label{C2}
    \textbf{C2: Beyond the Big-5}
\end{quote}

While earlier lexical studies of the English language revealed five core dimensions of personality, more recent analysis of both English and non-English languages (e.g., Italian, Dutch, German, Korean), based on larger sets of adjectives, has suggested the existence of a sixth core dimension: Honesty/Humility, giving rise to a new “Big-6” taxonomy (a.k.a. HEXACO)~\citep{ashton2007empirical}. Therefore, we encourage TPC researchers to  explore alternatives to the Big-5, in timely accordance with developments in personality psychology. For a comprehensive overview of (other) personality taxonomies, see \citet{cervone2022personality}.

%% file: writing/5Measurement.tex
\section{Measurement Quality}
\label{sec_measurements}
Personality traits are latent, theoretical variables (a.k.a. social science constructs), which can not be directly or objectively observed. Examples are emotions, prejudice and political orientation.
Thus, personality traits are inherently difficult to measure. We can only approximate the true underlying personality trait scores from often noisy observations collected using personality instruments such as self- and other-report questionnaires. Due to the uncertainty in this approximation process, measurements for personality traits likely contain non-negligible error. This error is called \textbf{measurement error}, defined as the difference between an observed measurement and its true value~\citep{dodge2003oxford}. 

The presence of error in personality measurements can have negative consequences for TPC research. For instance, consider the case of having substantial measurement error in questionnaire-based personality scores. When a TPC model treats these measurements as the gold standard for training and validation, the measurement error will likely propagate to the predictions, rendering the model less helpful (or even harmful) especially for diagnostic or clinical purposes. The study by \citet{Akrami2019AutomaticEO} lends support to this hypothesis, where the authors find TPC models to perform better on small datasets with low measurement error than on large datasets with high measurement error. Therefore, it is important that TPC researchers are aware of the presence and influence of measurement error and can deal with it. Unfortunately, none of the 60 TPC studies that we survey touch upon this issue, suggesting that this is an under-explored issue in TPC research. This observation inspires the next four challenges (3-6).





\begin{quote}
    \label{C3}
    \textbf{C3: Choose high-quality instruments}
\end{quote}

Collecting high-quality personality measurements begins with using high-quality instruments, be they questionnaires or models.\footnote{We consider prediction models and dictionaries/lexicons as personality instruments, just like questionnaires.} By high quality, we specifically mean high measurement quality. To determine the measurement quality of an instrument, it is important to understand the two components of measurement error: random error and systematic error, and how they relate to two quality criteria: reliability and validity. 

\textbf{Random error} refers to random variations in measurements across comparable conditions, due to factors that cannot be controlled~\citep{Trochim2015ResearchMT}. For instance, when someone completes a personality questionnaire twice, the responses may differ between the two attempts because the person misread a question in the second attempt. Random error is always present and unpredictable; it can be reduced but not eliminated. In NLP systems, random error can be due to random data splitting~\citep{gorman2019we}, stochastic algorithms~\citep{zhou2020TheCO}, and certain random processes in data annotations, such as sampling of annotators and random annotation mistakes~\citep{uma2021learning}. 

In contrast, \textbf{systematic error} occurs due to factors inherent to an instrument~\citep{Trochim2015ResearchMT}. For instance, a poorly constructed Big-5 questionnaire may contain an item that is used to measure neuroticism while in fact it does not. Consequently, anyone taking this questionnaire will get a biased estimate of their neuroticism. Thus, systematic error is foreseeable and often constant or proportional to the true value. As long as its cause is identified, systematic error can be removed. In NLP systems, systematic error can occur when spurious correlations (or “short-cuts”, instead of causal relationships) are learned~\citep{Wang2022IdentifyingAM}. 

\textbf{Reliability} and \textbf{validity} are the two criteria to the measurement quality of an instrument. The former concerns the extent to which an instrument can obtain the same measurement under comparable conditions, while the latter concerns the extent to which an instrument captures what it is supposed to~\citep{Trochim2015ResearchMT}. Random error reduces an instrument's reliability, while systematic error undermines its validity. Therefore, a high-quality instrument is a reliable and valid one. We describe below how we can find out about the validity and reliability of a personality instrument.

For personality questionnaires, especially of Big-5, it is relatively easy to determine their measurement quality because many corresponding validity and reliability studies exist (see \citealt{VANDERLINDEN2010315} for an overview). For model-based personality instruments, however, they rarely undergo comprehensive analysis of measurement quality. Typically, studies report the predictive performance of a personality model on some test data (using metrics like accuracy, recall, precision, F1, mean squared error, correlation). Such performance numbers can offer insight into the model's validity, assuming that the gold-standard personality measurements are low in error. However, the model’s validity in the presence of substantial measurement error in the data, as well as the model's reliability, remains unclear. Therefore, we urge future researchers to also examine and report both the validity and reliability of a model-based personality instrument. In Challenge 5, we discuss how this can be done.


However, it is important to exercise caution when selecting a personality instrument based on validity and reliability information obtained from earlier studies, as these studies had limitations in terms of the populations they examined (especially concerning demographic and linguistic characteristics), the time frames and contexts in which they were conducted. In a new study, the population of interest, the time and context may differ from those of previous studies. Consequently, researchers must carefully evaluate the validity and reliability evidence of an existing instrument in light of the specifics of the new study. 


Once a personality instrument is selected, it is also important to cite the source of the instrument and report its validity and reliability information. Among the 60 reviewed papers, 5 use self-identified outcomes (like one’s MBTI type or Big-5 scores mentioned in a tweet or user profile) where tracing down the instruments is impossible; 14 make use of proprietary instruments whose reliability and validity information is inaccessible to the public; among the 41 that use an existing personality instrument, 9 do not mention the specific instrument and only 4 report validity or reliability information based on previous studies. 

\begin{quote}
    \label{C4}
    \textbf{C4: Further reduce measurement error by study design}
\end{quote}

Even when the best possible instrument is used, there can still be substantial measurement error that results from other design factors of a study, especially when questionnaires are used. Factors like questionnaire characteristics (e.g., the number of questions, visual layout, topics, wording) and data collection modes (e.g., online, in person) can affect the measurement quality of the responses~\cite{biemer2013measurement}. Furthermore, factors related to respondents (e.g., inattention) can also affect measurement quality ~\citep{Fleischer2015InattentiveRI}. 



Therefore, when planning personality data collection using questionnaires, it can be beneficial to take into account different possible sources of measurement error. This helps to further reduce measurement error in questionnaire responses, in addition to using a valid, reliable questionnaire. For a comprehensive overview of factors that can influence measurement quality in questionnaires and the possible ways to control for them, we refer you to \citet{callegaro2015web} and \citet{biemer2013measurement}. We also encourage collaboration with survey methodology experts.



\begin{quote}
    \label{C5}
    \textbf{C5: Quantify measurement error}
\end{quote}

Now, assume that personality measurements have been collected.
The next step is to quantify both random and systematic error. 

For questionnaire-based measurements, we recommend using factor analysis, which is a type of latent variable model that relates a set of observed variables (e.g., personality questionnaire items) to some latent variables (e.g., one's true, underlying personality traits)~\citep{oberski2016mixture}. Depending on the model specification and data characteristics, factor analysis can decompose the total variation in the observed variables into different sources: variation due to the latent personality traits, variation due to systematic factors like questionnaire characteristics and the time of data collection, and finally, the unexplained variation (at the item level and at the questionnaire level). Larger variation due to the underlying personality traits and lower variation due to systematic factors are desirable, because they indicate higher measurement validity (i.e., less systematic error). In contrast, more unexplained variation indicates more random error (i.e., lack of reliability). Based on this variance decomposition, estimators of reliability and validity can be derived. We refer you to \citet[][Chapter 9-12]{saris2014design} for an overview of various factor analysis strategies and estimators of reliability and validity.

For model-based measurements, the number of measurements per personality trait and person is typically limited to one. Factor analysis cannot be applied to such data because the factor model is mathematically non-identifiable (i.e., there is no unique solution). Therefore, different methods for quantifying random and systematic error (or equivalently, reliability and validity) are needed. 

Random error is due to factors that cannot be controlled. Therefore, by varying the instrument or measurement condition along such factors, we can quantify the associated random error. In TPC models, one such factor is small variation in text (e.g., the use of singular vs. plural noun, which has not been linked to personality traits by prior research) that should not affect predictions. By introducing simple perturbations to the data, and comparing the new predictions with the ones based on the original data, we can gauge the degree of random error associated with this factor. This approach is analogous to \citet{Ribeiro2020BeyondAB}'s invariance tests. \citet{du2021assessing} also showcase such reliability analyses for word embedding-based gender bias scores. 

To quantify validity, apart from the usual performance metrics, we can check whether the predicted scores of different personality traits correlate with one another in expected ways. For instance, while the overall correlations should be low, as different personality traits are distinct constructs, some correlations should be more positive (e.g., between conscientiousness and agreeableness) than others (e.g., between openness and neuroticism). 
\citet{VANDERLINDEN2010315} provide an overview of empirical correlations between personality traits across demographic groups that can be expected.
To the best of our knowledge, no TPC study has assessed the correlations among predicted personality scores. 
Furthermore, it can be helpful to check whether the predicted personality scores relate to other constructs like emotions (if data is available) in expected ways. Such tests are conceptually similar to convergent and discriminant validity analyses in the social sciences~\citep{stachl2020personality}. Even in the presence of large measurement error in the data, they can be useful. For a more in-depth discussion of validity testing in machine learning and NLP, see \citet{jacobs2021measurement} and \citet{Fang2022EvaluatingTC}. 

\begin{quote}
    \label{C6}
    \textbf{C6: Correct for measurement error}
\end{quote}

Choosing high-quality personality instruments and reducing measurement error by design are likely the most important and effective ways to ensure high quality measurements. 
Once personality measurements have been collected, however, much less can be done about measurement error.

For questionnaire-based measurements, it might still be helpful to take a closer look at the results from factor analysis. For instance, do the measurements fit the assumed personality model (e.g., the Big-5)? If not, is it due to one or more questionnaire items that show unexpected relationships with the personality traits (e.g., the relationship is zero; the item correlates strongly with a different personality trait than expected)? If so, removing those problematic items can improve the validity of the personality measurements.  Are there items with large unexplained variances? If so, removing them may increase reliability. 

As for model-based measurements, if the personality models are proprietary or cannot be modified and retrained (e.g., due to lack of data or model details), then no correction for measurement error is possible. If retraining the model is possible, several techniques may help (see C12, C13 in \S\ref{sec_modelling}).

%% file: writing/6Datasets.tex
\section{Datasets}
\label{sec_datasets}
Across the 60 TPC studies, we find 41 unique datasets, which vary in terms of the personality taxonomy, instrument, type of text data, sample size, sample characteristics etc. Among them, however, only 18 are potentially accessible to other researchers (see Appendix~\ref{sec:appendixB}). 
Shareable datasets are key to advancing TPC research, as it leads to accumulation of data and allows for replication studies. This invites the next challenge:


\begin{quote}
    \label{C7}
    \textbf{C7: Construct shareable datasets}
\end{quote}


 

One obstacle to sharing TPC datasets is privacy preservation, as TPC datasets often contain identifiable information (e.g., names, locations, events) about data subjects. 
For instance, with social media posts, their authors can be easily found by using the content of the posts as search terms \citep{norman2022scraping}. We suggest two ways to make data sharing more privacy-preserving.


First,  data pseudonymization and anonymization techniques can be used. With pseudonymization, the data subjects can still be identified if additional information is provided. With anonymization, however, re-identification is impossible. Whether to pseudonymize or anonymize depends on many factors, such as the difficulty in data anonymization and the severity of re-identification. Nevertheless, for TPC datasets containing social media posts, anonymization is likely impossible. We refer you to \citet{Lison2021AnonymisationMF} for more information. 


Second, we can replace texts with paraphrases or synthetic data. The latter aims to “preserve the overall properties and characteristics of the original data without revealing information about actual individual data samples”~\citep{hittmeir2019utility}. However, whether these strategies are effective enough remain an open research question in NLP.


\begin{quote}
    \label{C8}
    \textbf{C8: Finer-grained measurements}
\end{quote}

All the 18 shareable datasets we find include only aggregated measurements of personality traits. Namely, for MBTI, only the classification types (e.g., INFP and ESTJ; instead of scores on each questionnaire item) are available; for Big-5, only the aggregated scores (e.g., means across items) for the five dimensions. This makes it impossible for other researchers to investigate measurement quality or train TPC models on the facet or item level. Even worse, some datasets provide no information about the personality instrument used. This especially concerns datasets that obtain personality labels from Twitter or Reddit based on the mention of MBTI or Big-5 information in a post or user profile (e.g., “INTJ”; “As an extravert…”). 

Other problematic treatments of aggregated personality measurements include further discretization, within-sample standardization or normalization to the target population. All leads to loss of information and limits the reusability of the measurements. 
Therefore, we suggest providing raw personality measurements, ideally on the item level.

\begin{quote}
    \label{C9}
    \textbf{C9: Include demographic information}
\end{quote}

We argue that the inclusion of demographic information (e.g., age, gender, education) can be important. Not only can this help researchers decide the appropriate personality instrument to use (in relation to the population of interest; see earlier discussion in \S\ref{sec_measurements}: C3), it can also provide additional useful features for TPC models. Furthermore, researchers can make use of the demographic information to diagnose the model (e.g., whether the measurement quality or the model’s prediction performance differs across demographic groups) (i.e., fairness). However, it is important to weigh the gain from including extra personal information against potential harm (see \S\ref{sec_ethics}: C14).


%% file: writing/7Evaluation.tex
\section{Performance Evaluation}
\label{sec_evaluation}
Across the 60 surveyed TPC papers, we identify two challenges related to performance evaluation:

\begin{quote}
    \label{C10}
    \textbf{C10: Use more appropriate and consistent performance metrics}
\end{quote}

11 out of the 60 studies model TPC as a regression task. Among them, 9 use Pearson's correlation between predicted personality trait scores and the true scores as the performance metric. However, correlation-based metrics can be misleading, as they register only ranks and do not reflect how accurate the predictions are on the original scale of the personality scores~\citep{stachl2020personality}. 

Some studies report mean squared error (MSE), which is arguably better than correlations because it quantifies the absolute difference between the predictions and the true values. However, MSE scores depend on the scale of the personality measurements and are not bounded, making interpretation and comparison (between studies) difficult. 




\citet{stachl2020personality} propose a better performance metric: $R^2=1-RSS/TSS$, where $RSS$ refers to the sum of squares of residuals and $TSS$ the total sum of squares. $R^2$ has several benefits. First, it can be considered a normalised version of MSE, which has an upper limit of 1 (perfect agreement). Second, $R^2$ has a natural zero point, which occurs when the mean is used as the prediction. Third, when the model makes worse predictions than a simple mean baseline, $R^2$ becomes negative.

While two studies report $R^2$, their calculation of $R^2$ is unclear. The researchers may have calculated $R^2$ not based on the formula shown earlier, but by squaring Pearson's correlations. This would lead to always positive $R^2$, which can be misleading. 

In the 33 studies that model TPC as a classification task, a more diverse set of metrics are used (i.e., accuracy, recall, precision, F1, and AUC, in macro, micro, weighted or unweighted forms). One problem, however, is that different studies report different metrics (sometimes, only one). This makes comparison across studies difficult. We encourage future researchers to report all common metrics for classification studies (like those mentioned above). 





\begin{quote}
    \label{C11}
    \textbf{C11: Report performance expectation}
\end{quote}

While it is normal to optimize the prediction performance of TPC models, it is also important to set correct expectations: What kind of performance can we realistically expect? How accurate can personality predictions be when only (short) text is used? How good does the performance need to be for a particular system? \citet{stajner_why_2021} make the first question even more relevant, as they find either few or mixed MBTI-related signals in typical text data used for MBTI prediction. 

Quantifying measurement error in the personality scores used for modelling can help researchers to set clearer expectations about model performance, because high systematic error will limit the model's generalizability to new data, while high random error will result in unstable predictions. 


Thus, setting expectations "forces" researchers to learn more about their data and to avoid unrealistic expectations that may lead to problematic research practices like cherry-picking results. Unfortunately, none of the 60 reviewed papers discusses performance expectation.

%% file: writing/8Models.tex
\section{Modelling Choices}
\label{sec_modelling}
Most TPC studies model different personality traits (i.e., dimensions) separately. This strategy is undesirable, because it ignores the correlations among personality traits that models can learn from. Modelling personality traits jointly may also help to prevent overfitting to a specific trait and thus learn more universal personality representations~\citep{Liu2019MultiTaskDN}. Hence, the next challenge:

\begin{quote}
    \label{C12}
    \textbf{C12: Joint personality modelling}
\end{quote}

Out of the 60 studies, only 5 attempt at (some form of) joint modelling of personality traits. \citet{yang_learning_2021} implement a transformer-based model to predict MBTI types, where the use of questionnaire texts allows the model to infer automatically the relevant MBTI dimension, and hence removes the need for independent modelling of different MBTI dimensions. \citet{gjurkovic_reddit_2018}, \citet{bassignana_matching_2020} and \citet{hosseinia_usefulness_2021} frame the prediction of MBTI types as a 16-class classification task (as there are in total 16 MBTI types), thereby using only one single model. \citet{hull_personality_2021} apply “stacked single target chains”~\citep{Xioufis2016MultitargetRV}, which feeds the predictions of one personality trait back in as features for the prediction of the next trait(s). 

Multitask learning may also be useful, which trains a model on multiple tasks simultaneously and thus might help to improve the generalizability of the model~\citep{caruana1997multitask}. In addition, \citet{stachl2020personality} suggest modifying a model’s loss function such that the correlations between theoretically distinct constructs are minimised. Building on this idea, we can also specify the loss function in a way that it not only focuses on general prediction performance but also minimises the difference between the predicted covariance matrix and the observed covariance matrix of personality traits. 

\begin{quote}
    \label{C13}
    \textbf{C13: Build on best modelling choices}
\end{quote}

As the field progresses, it is important to not only investigate new modelling ideas, but also accumulate knowledge about best modelling practices. We list below several empirically supported ideas that should not slip past the community’s attention. 

First, researchers should leverage the texts in personality questionnaires. \citet{kreuter_items_2022}, \citet{vu_predicting_2020} and \citet{yang_learning_2021} find that incorporating personality questionnaire texts into model learning can lead to better personality predictions.

Second, when sample sizes are small, data augmentation and dimensionality reduction techniques are beneficial. \citet{kreuter_items_2022} show that using data augmentation to increase the training size of personality questionnaire items leads to better predictions. \citet{v_ganesan_empirical_2021} show that PCA helps to overcome the problem of fine-tuning large language models with a small TPC dataset.

Third, incorporating personality-related variables (e.g., psycholinguistic features, emotions, user interests, demographics, opinions and brand preference) as features, if available, can improve personality predictions \citep{kerz_pushing_2022, cornelisse_inferring_2020, kishima_construction_2021, li_prompt-based_2022, hosseinia_usefulness_2021, yang_using_2015}.

%% file: writing/10Ethics.tex
\section{Ethics and Fairness}
\label{sec_ethics}
Out of the 60 reviewed papers and the 2 additional survey papers, only 7 provide some reflection about ethics and none about fairness. This can be because ethics and fairness only became central in NLP recently. 
Nevertheless, the last two challenges:

\begin{quote}
    \label{C14}
    \textbf{C14: More ethical and useful TPC}
\end{quote}

As useful as TPC can be, it is important to ask whether gathering personal information like personality or computing them is necessary. This is especially relevant for research where TPC is only an intermediate step to another end such as opinion mining~\citep{hosseinia_usefulness_2021}, dialogue generation~\citep{mairesse_trainable_2008} and brand preference prediction~\citep{yang_using_2015}. Such studies typically argue that the computed personality traits can be used as features for another task and that it leads to better task performance; however, they do not consider alternatives (e.g., replacing the prediction of personality traits with using lexical cues that are non-personal but still indicative of personality). Thus, we encourage researchers to justify PC and to find alternatives when PC is only a means.

Even when TPC can be justified, it is important to reduce potential harm. For instance, many TPC studies and datasets make use of public social media profiles for predicting personality traits. While this is often legal, no explicit consent for PC is obtained from the social media users, which makes using public social media data an ethically ambiguous issue~\citep{norman2022scraping}. \citet{Boeschoten2022AFF} proposes a privacy-preserving data donation framework that may help to alleviate this problem, where data subjects can voluntarily donate their data download packages (e.g., from social media accounts) for research and give explicit consent.

To further increase the benefit of TPC, we can consider applying it to clinical, professional or educational settings, where (traditional) personality assessment has proven useful and relevant (e.g., personalised treatments; career recommendation; individualised learning). None of the 60 TPC studies in our review investigates these applications.

\begin{quote}
    \label{C15}
    \textbf{C15: Research on Fair TPC}
\end{quote}

Fairness research in machine learning concerns identifying and mitigating biases that may be present within a system, particularly towards specific groups~\citep{mehrabi_survey_2021}. For instance, a fair TPC model should exhibit equal predictive performance across different demographic groups. 
Among many others, biased training data is a significant factor contributing to algorithmic bias. From the perspective of measurement quality, the choice of a personality instrument that lacks equal validity and reliability across all demographic groups of interest can introduce variations in the quality of personality measurements among different groups. Consequently, these discrepancies can perpetuate algorithmic bias within the system.

Remarkably, none of the 60 TPC papers we survey address the topic of fairness. Therefore, there is a clear need for future research on fairness in the context of TPC.

%% file: writing/11Conclusion.tex
\section{Conclusion}

In this paper, we review 60 TPC papers from the ACL Anthology and identify 15 challenges that we consider deserving the attention of the research community. We focus on the following 6 topics: personality taxonomies, measurement quality, datasets, performance evaluation, modelling choices, as well as ethics and fairness. While some of these topics (e.g., personality taxonomies and ethics) have been discussed elsewhere, we provide new perspectives. Furthermore, in light of these challenges, we offer concrete recommendations for future TPC research, which we summarise below:

\begin{itemize}
    \item Personality taxonomies: Choose Big-5 over MBTI; Try modelling facets and using other taxonomies like HEXACO where appropriate. 
    \item Measurement quality: Pay attention to measurement error in personality measurements, be they based on questionnaires or models; Try to reduce measurement error by design (e.g., choose higher-quality instruments; use better data collection practices); Provide quality evaluation (i.e., validity and reliability) for any new (and also existing) approaches.
    \item Datasets: Make TPC datasets shareable, which should also contain fine-grained personality measurements and descriptions of the target population;
    \item Performance evaluation: Report a diverse set of performance metrics; Report $R^2$ for a regression task. 
    \item Modelling choices: Make use of their psychometric properties when modelling personality traits (e.g., use joint modelling; modify the loss function to preserve the covariance information); For even better predictions, try incorporating personality questionnaire texts, applying data augmentation and dimensionality reduction techniques, as well as incorporating more personality-related variables. 
    \item Ethics and fairness: Avoid unnecessary TPC; Apply TPC to clinical, professional and educational settings; Investigate fairness.
    \item Lastly, engage in (interdisciplinary) research work with survey methodologists, psychologists, and psychometricians.
\end{itemize}

We hope that our paper will inspire better TPC research and new research directions.

\section*{Limitations}
Our paper has some limitations. First, we do not give detailed instructions about techniques that we recommended (e.g., factor analysis, synthetic data generation). We rely on our readers’ autonomy to acquire the necessary information (that is specific to their research projects) by further reading our recommended references. Second, we only survey TPC papers included in the ACL Anthology, despite other TPC papers existing outside this venue. While this means that the challenges we identified might be specific to these papers, we believe they are still a good representation of the TPC research done in NLP. Lastly, we limit our discussion to text data. It would be beneficial for future research to also discuss challenges facing PCs based on other types of data (e.g., images, behaviours, videos), which may offer additional insights to TPC research.

\section*{Statement of Ethics and Impact}
Our work provides a critical evaluation of past TPC research, where we present 15 open challenges that we consider deserving the attention of the research community. For each of these challenges, we offer concrete suggestions, thereby hoping to inspire higher-quality TPC research (e.g., more valid and reliability personality measurements, better datasets, better modelling practices). We also discuss issues related to ethics and fairness (see \S\ref{sec_ethics}). We hope to see more ethical and fair TPC research.

\section*{Acknowledgements}
We thank Dr. Dong Nguyen, Anna Wegmann, Christian Fang and Yupei Du from Utrecht University, as well as our anonymous reviewers for their valuable feedback on this paper.  
This work is partially supported by the Dutch Research Council (NWO) (grant number: VI.Vidi.195.152 to D. L. Oberski).